\documentclass{IEEE-Con-Sys-mag}

\usepackage[compress]{cite}

\usepackage{bm}
\usepackage{derivative}

\newcommand{\R}{\mathbb{R}}

\newcommand{\bx}{\mathbf{x}}
\renewcommand{\bf}{\mathbf{f}}
\newcommand{\bg}{\mathbf{g}}
\newcommand{\bk}{\mathbf{k}}
\newcommand{\bu}{\mathbf{u}}

\newcommand{\bp}{\mathbf{p}}
\newcommand{\bn}{\mathbf{n}}
\newcommand{\bq}{\mathbf{q}}
\newcommand{\bv}{\mathbf{v}}
\newcommand{\bomega}{\boldsymbol{\omega}}
\newcommand{\be}{\mathbf{e}}
\newcommand{\bzero}{\mathbf{0}}
\newcommand{\bJ}{\mathbf{J}}
\newcommand{\bR}{\mathbf{R}}
\newcommand{\bM}{\mathbf{M}}

\jname{IEEE CONTROL SYSTEMS}
\pubyear{2026}

\begin{document}

\title{Safety Guardrails \break in the Sky\stitle{Realizing Control Barrier Functions on the VISTA F-16 Jet}}

\author{
Andrew W. Singletary,
Max H. Cohen,
Tamas G. Molnar, and
Aaron D. Ames
}
\affil{
}

\maketitle



\chapterinitial{T}he deployment of autonomous systems requires the satisfaction of strict safety specifications that ensure these systems do not harm themselves or their environments.
These safety specifications may encompass a variety of constraints for different systems, including limits on the configurations, velocities, and torques of robotic systems, collision avoidance constraints for ground vehicles, or flight envelope bounds for aircraft, to name a few examples.
To achieve end-to-end safety, functional safety systems have been utilized and demonstrated success, allowing autonomous behavior under normal (safe) operating conditions but intervening when a safety limit is tripped.
During intervention, these systems often use a fail-safe mechanism by switching over to a backup safety maneuver that controls the system in a safe, but potentially conservative way, often sacrificing the completion of tasks and the fulfillment of performance objectives in the interest of safety.

As autonomous systems become more sophisticated, there is an increasing need for executing highly dynamic high-performance maneuvers on the edge of the system's safe operating domain.
Examples include high-speed collision-free driving on autonomous vehicles, dexterous locomotion and manipulation on robotic systems, or highly dynamic maneuvers on unmanned aerial vehicles.
With these requirements, conservative safety trips that prevent the achievement of performance objectives may no longer be satisfactory as they may be overly restrictive and prevent autonomous systems from achieving their goals.
This creates a need for more dynamic notions of safety and for runtime assurance modules that allow the system operate on the edge of its safety envelope but not beyond.
Importantly, such dynamic safety modules should come with formal certificates of safe behavior while remaining minimally invasive, allowing nominal autonomous operations to continue when they are safe and gradually taking control authority when approaching the safety limit.

In this paper, we introduce {\em Guardrails} --- a novel framework to runtime assurance and safety filtering for highly dynamic autonomous and semi-autonomous systems. This is achieved through a unique control strategy that blends nominal commands, potentially coming from a human operator or AI agent, with safe control actions in a smooth, non-disruptive fashion, with formal guarantees of safe behavior backed by the theory of control barrier functions (CBFs). While Guardrails applies to general autonomous systems, such as humanoids and quadrotors, our emphasis in this paper is on the application of Guardrails to safe \emph{flight control}, an arguably more safety-critical setting than those found in traditional robotics applications. Here, we illustrate how integrating Guardrails into an existing flight control system on a fixed-wing aircraft enables pilots to operate safely, even on the edge of the flight envelope. We demonstrate how Guardrails' blended controllers supervise both human pilots and AI-based flight controllers to maintain a variety of safety constraints, including g-limits, altitude ceiling and floor limits, geofence constraints, and the combination thereof.
Importantly, Guardrails modifies pilot authority only when approaching a constraint limit and has no impact on flight characteristics otherwise, making it a minimally invasive safety system. As an add-on safety module, Guardrails enables the safe deployment of advanced autonomy on fly-by-wire vehicles.

In collaboration with the US Air Force Test Pilot School, we implemented and experimentally tested Guardrails on the Variable In-Flight Simulation Test Aircraft (VISTA), a modified F-16 fighter jet. A total of fourteen real-life test scenarios were executed during five flights at the Edwards Air Force Base, where Guardrails was used to supervise a human pilot aboard the VISTA and ensure the satisfaction of various operational constraints. To the best of our knowledge, this marks the first implementation of a CBF-based control strategy on a full-scale fixed-wing aircraft. In all tests, Guardrails ensured the satisfaction of the prescribed safety constraints, even in the presence of adversarial pilot inputs that were deliberately intended to violate these constraints.

In what follows, we frame Guardrails in the context of the broader literature, cover the technical background of our approach, the results of our flight tests, and discuss the capabilities and limitations of Guardrails.
Finally, we discuss potential future directions to extend Guardrails towards a comprehensive runtime assurance system that guards safety in the sky --- protecting both the pilots and passengers, the aircraft, and their environment. We also highlight Guardrail's broader implications for safe autonomous systems. 


\begin{summary}
\summaryinitial{T}he advancement of autonomous systems --- from legged robots to self-driving vehicles and aircraft --- necessitates executing increasingly high-performance and dynamic motions without ever putting the system or its environment in harm's way.
In this paper, we introduce {\em Guardrails} --- a novel runtime assurance mechanism that guarantees \emph{dynamic safety} for autonomous systems, allowing them to safely evolve on the edge of their operational domains.
Rooted in the theory of control barrier functions,
Guardrails offers a control strategy that carefully blends commands from a human or AI operator with safe control actions to guarantee safe behavior.
To demonstrate its capabilities, we implemented Guardrails on an F-16 fighter jet and conducted flight tests where Guardrails supervised a human pilot to enforce g-limits, altitude bounds, geofence constraints, and combinations thereof.
Throughout extensive flight testing, Guardrails successfully ensured safety, keeping the pilot in control when safe to do so and minimally modifying unsafe pilot inputs otherwise.
\end{summary}

\subsection{Related Works}
We now provide a brief overview of the literature on collision avoidance systems for aircraft and safety-critical control in a broader context, and we position our work relative to the state of the art.

\subsubsection{Collision Avoidance and Run-Time Assurance for Aircraft}


Existing flight control systems offer various collision avoidance features that monitor safety and intervene when necessary~\cite{guan2020survey}.
Automatic ground collision avoidance systems  (GCAS) prevent controlled flight into terrain accidents where the aircraft crashes into the ground~\cite{burns2011autogcas, swihart2011autogcas, suplisson2015optimal, heidlauf2018verification, carpenter2019autogcas, kirkendoll2021}.
Being the leading cause of F-16 fatalities, it is especially important for military jets to prevent ground collisions, because  maneuvers with large g-forces may cause pilots to lose consciousness or become disoriented.
GCAS monitors the terrain and the aircraft trajectory, and when a collision is predicted, it commands a maneuver that rolls the aircraft to wing level and pulls it up.
In a similar fashion, automatic airborne collision avoidance systems (ACAS) serve to prevent air-to-air collisions between two aircraft~\cite{williams2004airborne, chamlou2009future, kochenderfer2012nextgeneration}.
ACAS supplements air traffic control and operates without the use of ground-based equipment.
This system monitors the airspace around an aircraft and when other aircraft are detected with the risk of mid-air collision, ACAS warns the pilot and recommends a maneuver to mitigate the risk.
ACAS systems have been studied extensively in the context of verification~\cite{tang2014casual, vonessen2014analyzing, jeannin2015formal, lee2015adaptive}, optimization~\cite{kochenderfer2015optimized}, implementation on unmanned aircraft~\cite{manfredi2016introduction, owen2019acasxu, deaton2020evaluating}, security~\cite{smith2022understanding}, and advancements through artificial intelligence~\cite{christensen2024advancing}.
The most common implementation of ACAS, called the automatic traffic collision avoidance system (TCAS)~\cite{harman1989tcas, williamson1989development, livadas2000highlevel, kuchar2007traffic, murugan2010tcas, munoz2013tcas, tang2017review, tang2018casual, longo2024onacollision}, is utilized on commercial aircraft.
Finally, ACAS and GCAS may be fused into automatic integrated collision avoidance systems (ICAS), preventing both ground and airborne collisions. More recently, such ideas have been extended to encompass ground and airborne collisions with multiple aircraft as well as fixed obstacles in the environment that may represent, e.g., no-fly zones or bad-weather areas \cite{Corraro2022}.

Many of these collision avoidance systems act as a supervisor that lets the pilot operate without interruption in safe situations and intervenes when safety is in danger of violation.
This operational principle, often referred to as runtime assurance (RTA)~\cite{fuller2020rta, Nagarajan2021, hobbs2023rta}, is able to supervise not only human pilots but also complex flight controllers for which safety is difficult to verify.
This is crucial for facilitating safe autonomy, especially with the increasing presence of artificial intelligence in flight systems \cite{HobbsSciTech23}.
RTA has shown success in achieving safety, for example, in unmanned aircraft navigation along corridors~\cite{schierman2020rta}, aerial refueling tasks~\cite{costello2023rta}, and neural network-based aircraft taxiing~\cite{cofer2020rta}.


Some existing RTA mechanisms
are disruptive in nature, taking the authority away from the pilot or the nominal flight controller in dangerous situations and fully switching over to predefined safe backup maneuvers.
Rather than using disruptive switching RTA, recently the attention has shifted towards optimization-based RTA~\cite{hobbs2023rta} --- continuously active control systems that take into account what the human operator or the primary flight controller commands and optimize for the closest safe action to minimize the intervention.
In a similar fashion, Guardrails also offers a  strategy to blend primary commands with safe actions, but in a less computationally intensive  optimization-free manner while preserving partial control authority for the human or AI operator.



\subsubsection{Safety-Critical Control}
Motivated by the increasing levels of autonomy of modern engineering systems, the past decade has witnessed a surge of research in the area of \emph{safety-critical control} in which safety --- a system-theoretic property formalized through the framework of set invariance \cite{Blanchini} --- is a top design priority. Such control designs are often carried out using a class of controllers that have become known as \emph{safety filters}. These controllers supervise a nominal (typically performance-based) controller and adjust its actions so that the resulting closed-loop system is safe \cite{AmesCSM23,FisacARCRAS23}. In recent years, safety filters have been constructed using various control methodologies including control barrier functions (CBFs)~\cite{AmesTAC17, GurrietICCPS18,AmesECC19}, model predictive control \cite{MayneRawlingsDiehl,BorelliBemporadMorari}, Hamilton-Jacobi reachability \cite{TomlinTAC05,HerbertCDC21,ChoiArXiV24}, and reference governors \cite{KolmanovskyAutomatica17}. Each of these approaches comes with their advantages and disadvantages, with more details outlined in comprehensive surveys such as \cite{AmesCSM23,FisacARCRAS23}. 

As noted earlier, our Guardrails framework is based on the theory of CBFs, but operates in a somewhat untraditional manner compared to most CBF approaches. In general, CBF-based safety filters are instantiated as convex optimization problems, typically a quadratic program, that can be solved efficiently at any state to produce a safe control input~\cite{AmesTAC17, GurrietICCPS18,AmesECC19}. Our Guardrails framework differs from these traditional CBF approaches in two ways: i) the computation of the CBF itself and ii) the structure of the safety filter. To compute the CBF, we leverage the framework of \emph{backup CBFs}~\cite{gurriet2020scalable}. At its core, this method takes in a known small control invariant set, which can be constructed offline, that is expanded to a larger control invariant set online via rollouts of the system trajectory under a pre-specified ``backup" control policy. The end result is an \emph{implicit} CBF that ensures compatibility with hard input bounds, a property that is challenging to ensure using traditional CBF approaches. One of the main limitations of this approach is the need to compute the gradient of the resulting implicit CBF, which requires computing the sensitivity of the system's flow under the backup policy and scales poorly to higher dimensional systems, such as the aircraft models used in our results. Guardrails resolves this issue by using a blended safety filter that does not require gradients of the CBF, which facilitates scalability to higher dimensional systems. The efficacy of these blended safety filters has been previously illustrated through their application to high-speed geofencing with quadrotors~\cite{singletary2021comparative, singletary2022onboard, singletary2022safedrone}. Here, we extend this methodology to a more complex and safety-critical setting by deploying Guardrails on a fixed-wing aircraft. 

Outside of Guardrails, CBFs have shown success in ensuring safety on fixed-wing aircraft in the context of probabilistic safety certificates~\cite{Luo2019}, learning-based~\cite{Scukins2021} and data-driven CBFs~\cite{squires2021modelfree}, and multi-aircraft control~\cite{squires2022composition}.
Furthermore, high-order CBFs~\cite{Zhou2020}, barrier Lyapunov functions~\cite{xu2022barrierlyapunov}, and robust CBFs~\cite{zheng2023constrained} have also been used for trajectory tracking on fixed-wing aircraft.
Finally, our prior work has established simultaneous collision avoidance and geofencing on fixed-wing aircraft using an optimization-based RTA system with CBFs~\cite{molnar2025fixedwing}.
These works, however, are restricted to theoretical and simulation studies, and the literature has not yet reported hardware implementations of CBF-based aircraft RTA systems.

\subsection{Overview of Guardrails}

\begin{figure*}
    \centerline{\includegraphics[scale=1]{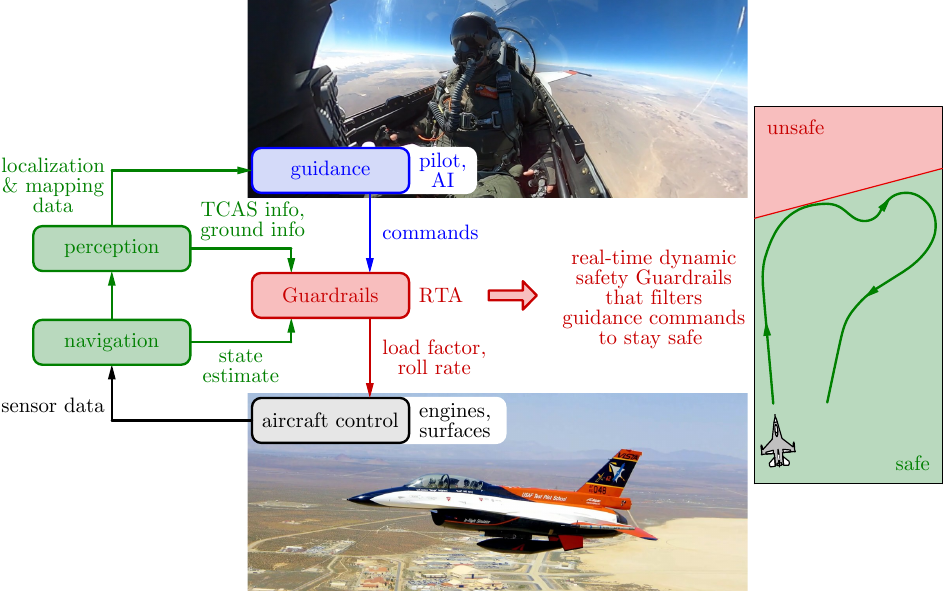}}
    \caption{Overview of Guardrails --- a runtime assurance system that supervises artificial intelligence or human pilots in real time for safe highly dynamic maneuvers.
    Guardrails was implemented on the X-62 Variable-stability In-flight Simulator Test Aircraft (VISTA) shown at the bottom, and evaluated during flight tests with the Edwards Air Force Base Test Pilot School.
    }
    \label{fig:guardrails}
\end{figure*}

Before presenting the technical details of our method and our experimental results, we first provide a high-level overview of our Guardrails framework. Guardrails is a module that integrates into a system's existing autonomy stack, such as the VISTA as outlined in Fig. \ref{fig:guardrails}, to ensure the safety of the resulting system. Guardrails acts as a \emph{safety filter} by adjusting nominal commands sent to the autonomy stack, such as those produced by a pilot, so that the eventual actions taken by the system are deemed to be safe. As Guardrails is based on the theory of CBFs, safety here is synonymous with set invariance: Guardrails computes control inputs that ensure that the trajectory of the system remains within a desirable region of the system's operating domain. To compute such inputs, Guardrails requires: i) a dynamical model of the system with state $\bx\in\mathcal{X}$ and input $\bu\in\mathcal{U}$; ii) a safety requirement, mathematically encoded by the nonnegativity of a
scalar safety function $h\,:\,\mathcal{X}\rightarrow\R$; and iii) a backup maneuver $\bk_{\rm b}\,:\,\mathcal{X}\rightarrow\mathcal{U}$ that brings the system to a nominal operating condition from a small set of initial conditions (e.g., returning to level flight).

Guardrails leverages these components to compute a safe operating region --- the set of all states where the system can safely return to a nominal operating condition without ever violating safety requirements --- and monitors the system's proximity to the boundary of this safe operating region. Based on this, Guardrails smoothly blends desired actions from a performance controller or a human operator with the actions from the backup maneuver via a scalar parameter $\lambda$ taking values between 0 and 1, with $\lambda=0$ allowing desired actions to freely pass through the autonomy stack, $\lambda=1$ overriding these desired actions in favor of those corresponding to the prespecified backup maneuver, and ${\lambda\in(0,1)}$ carefully blending these two actions to ensure safety of the overall system.

\section{Technical Overview}
In this section, we provide a more in-depth technical overview of the tools used within our Guardrails framework for ensuring dynamic safety of autonomous systems. 

\subsection{Safety-Critical Control}
Our Guardrails framework is based on the control-theoretic notion of set invariance \cite{Blanchini}, which, in recent years, has proven to be a powerful proxy for safety of autonomous systems \cite{AmesECC19}. In this setting, an autonomous system of interest is modeled as a nonlinear control system:
\begin{equation}\label{eq:dyn}
    \dot{\bx} = \bf(\bx) + \bg(\bx)\bu,
\end{equation}
where ${\bx\in\mathcal{X}\subset\R^n}$ is the system state with state space ${\mathcal{X}\subset\R^n}$ and ${\bu\in\mathcal{U}\subset\R^m}$ is the control input with admissible control space ${\mathcal{U}\subset\R^m}$. Here, ${\bf\,:\,\mathcal{X}\rightarrow\R^n}$ and ${\bg\,:\,\mathcal{X}\rightarrow\R^{n\times m}}$ are locally Lipschitz functions characterizing the system dynamics: $\bf$ captures the drift dynamics of the system and columns of $\bg$ model the control directions.

Safety of such control systems is often
centered on the notion of set invariance.
To introduce these ideas, let ${\bk\,:\,\mathcal{X}\rightarrow\mathcal{U}}$ be a locally Lipschitz feedback controller.
By taking ${\bu=\bk(\bx)}$ in~\eqref{eq:dyn}, this produces the closed-loop system:
\begin{equation}\label{eq:dyn-cl}
    \dot{\bx} =
    \bf(\bx) + \bg(\bx)\bk(\bx).
\end{equation}
Since both the dynamics and controller are locally Lipschitz, the closed-loop system is as well, and thus admits a unique continuously differentiable solution defined on a maximal interval of existence for each initial condition $\bx_0\in\mathcal{X}$.
We will denote by $\bm{\varphi}(t,\bx_0)$ the \emph{flow} associated with the closed-loop system in~\eqref{eq:dyn-cl}, that is, the state reached at time $t$ when starting from state $\bx_0$. The following definitions formalize the concept of safety for autonomous systems via set invariance.

\begin{definition}\label{def:forward-invariance}
    A set $\mathcal{S}\subset\mathcal{X}$ is said to be forward invariant for the closed-loop system in~\eqref{eq:dyn-cl} if for each $\bx_0\in\mathcal{S}$ we have $\bm{\varphi}(t,\bx_0)\in\mathcal{S}$ for all $t\geq0$. If $\mathcal{S}$ is forward invariant, the closed-loop system is said to be safe on $\mathcal{S}$.
\end{definition}

\begin{definition}\label{def:control-invariance}
     A set $\mathcal{S}\subset\mathcal{X}$ is said to be control invariant for the control system in~\eqref{eq:dyn} if there exists a locally Lipschitz feedback controller $\bk\,:\,\mathcal{X}\rightarrow\mathcal{U}$ such that $\mathcal{S}$ is forward invariant for the resulting closed-loop system in~\eqref{eq:dyn-cl}.
\end{definition}

Note that Def.~\ref{def:forward-invariance} applies to closed-loop systems --- those equipped with feedback controllers --- whereas Def.~\ref{def:control-invariance} is tailored to control systems where the input is yet to be specified as in~\eqref{eq:dyn}, and characterizes the existence of controllers that ensure the safety of the corresponding closed-loop system. In the following subsection, we illustrate how these fundamental concepts may be used to systematically design controllers enforcing safety of autonomous systems.

\subsection{Implicit Control Invariant Sets}
Perhaps the most fundamental challenge in safety-critical control is constructing control invariant sets. When the admissible control space $\mathcal{U}$ is unbounded, i.e., when ${\mathcal{U}=\R^m}$, there exist systematic approaches to construct these control invariant sets \cite{WeiTAC22,TanTAC22,CohenARC24}. When this control space is bounded $\mathcal{U}\subset\R^m$, however, constructing control invariant sets becomes quite difficult. In general, these sets are characterized as the zero superlevel set of the value function satisfying a particular Hamilton-Jacobi-Bellman equation \cite{TomlinTAC05}. While strides have been made to solve these partial differential equations more efficiently, one is ultimately limited by the ``curse of dimensionality," which, in practice, precludes one from applying these techniques to high-dimensional systems.

In this section, we present a technique that overcomes these limitations through the notion of an \emph{implicit} control invariant set, defined using the flow of the system.
To this end, let:
\begin{equation}\label{eq:constraint-set}
    \mathcal{C} = \{\bx\in\mathcal{X}\,:\,h(\bx) \geq 0\},
\end{equation}
denote a state constraint set that characterizes operational limits on the autonomous system, where $h\,:\,\mathcal{X}\rightarrow\R$ is a locally Lipschitz function. This set could, for example, describe hard limits on various components of the system state (e.g., altitude limits) or admissible regions of the state space in which the system must operate (e.g., within the boundaries of a geofence). While it may seem restrictive to encode a collection of complex constraints into the superlevel set of a single function, this function, unlike traditional CBF approaches, need only be Lipschitz continuous; it can thus be taken as, e.g., the signed distance to a set of failure states or as the Boolean combination of multiple constraint functions.

Although it is desired to keep the system within $\mathcal{C}$, it may be the case that $\mathcal{C}$ is not control invariant -- it may be \emph{impossible} to find a controller that ensures invariance of this set. In this situation, we assume the existence of a smaller control invariant set, termed the \emph{backup safe set}:
\begin{equation}\label{eq:backup-set}
    \mathcal{C}_{\rm b} = \{\bx\in\mathcal{X}\,:\,h_{\rm b}(\bx) \geq 0\}\subset\mathcal{C},
\end{equation}
where ${h_{\rm b}\,:\,\mathcal{X}\rightarrow \R}$ is a locally Lipschitz function. In particular, this set is assumed to be forward invariant for the closed-loop system:
\begin{equation}\label{eq:dyn-backup}
    \dot{\bx} =
    \bf(\bx) + \bg(\bx)\bk_{\rm b}(\bx),
\end{equation}
under the influence of a locally Lipschitz \emph{backup controller} ${\bk_{\rm b}\,:\,\mathcal{X}\rightarrow\mathcal{U}}$. This backup controller represents a fail-safe maneuver that brings the system into a safe operating condition.
With this backup controller in hand, we define the \emph{implicit safe set}:
\begin{equation}\label{eq:implicit-safe-set}
    \mathcal{S} = \{\bx\in\mathcal{X}\,:\,h_{I}(\bx) \geq 0 \}
\end{equation}
\begin{equation}\label{eq:implicit-h}
    h_{I}(\bx) = \min\left\{h_{\rm b}\left(\bm{\varphi}_{\rm b}(T,\bx)\right),\,\min_{\theta\in[0,T]}h\left(\bm{\varphi}_{\rm b}(\theta,\bx)\right) \right\},
\end{equation}
where $\bm{\varphi}_{\rm b}(\theta,\bx)$ is the flow of the closed-loop system in~\eqref{eq:dyn-backup} under the backup controller,
starting from state $\bx$ over time ${\theta \in [0,T]}$,
and ${T>0}$ is the backup horizon. The intuition behind the above definition is as follows: if $\bx\in\mathcal{S}$, then, by taking $\bu=\bk_{\rm b}(\bx)$, the trajectory of the closed-loop system remains within $\mathcal{C}$ for $T$ units of time and enters $\mathcal{C}_{\rm b}\subset\mathcal{C}$ in, at most, $T$ units of time. Since $\mathcal{C}_{\rm b}$ is forward invariant under $\bk_{\rm{b}}$ and $\mathcal{C}_{\rm b}\subset\mathcal{S}$, the system may remain in $\mathcal{C}_{\rm b}\subset\mathcal{S}$ for all time thereafter, implying that $\mathcal{S}$ is also forward invariant. The following theorem formally characterizes the properties of $S$ and $h_{I}$.

\begin{theorem}[backup set method~\cite{gurriet2020scalable}]\label{thm:implicit-safety}
    Consider the control system in~\eqref{eq:dyn}, a state constraint set $\mathcal{C}$ as in~\eqref{eq:constraint-set}, and a backup safe set $\mathcal{C}_{\rm b}\subset\mathcal{C}$ as in~\eqref{eq:backup-set}, assumed to be forward invariant for the closed-loop system in~\eqref{eq:dyn-backup} under a locally Lipschitz backup controller $\bk_{\rm b}\,:\,\mathcal{X}\rightarrow\mathcal{U}$. Provided that $h\,:\,\mathcal{X}\rightarrow\R$ and $h_{\rm b}\,:\,\mathcal{X}\rightarrow\R$ are locally Lipschitz continuous, then:
    \begin{enumerate}
        \item $h_{I}\,:\,\mathcal{X}\rightarrow\R$ as in~\eqref{eq:implicit-h} is locally Lipschitz continuous;
        \item $\mathcal{S}\subset\mathcal{C}$ as in~\eqref{eq:implicit-safe-set} is control invariant.
        \item $\mathcal{S}\subset\mathcal{C}$ as in~\eqref{eq:implicit-safe-set} is forward invariant for the closed-loop system under the backup controller.
    \end{enumerate}
\end{theorem}

\subsection{Blended Safety Filters}
In the previous subsection, we presented a framework for generating implicit control invariant sets. We now turn our attention to the construction of controllers that allow for maximal operational freedom while still ensuring safety, broadly characterized as \emph{safety filters}. Here, we leverage a blended safety filter from~\cite{singletary2022onboard} that continuously blends a desired controller ${\bk_{\rm{d}}\,:\,\mathcal{X}\times\R\rightarrow\mathcal{U}}$ (which may represent a human or AI operator) and a backup controller ${\bk_{\rm{b}}\,:\,\mathcal{X}\rightarrow\mathcal{U}}$ based on the value of $h_{I}(\bx)$ as:
\begin{equation}\label{eq:regulation-safety-filter}
    \bk(\bx,t) = \big( 1 - \lambda\left(h_{I}(\bx) \right) \big) \bk_{\rm{d}}(\bx,t) + \lambda\left(h_{I}(\bx) \right)\bk_{\rm{b}}(\bx),
\end{equation}
where ${\lambda\,:\,\R\rightarrow[0,1]}$ is any locally Lipschitz function satisfying ${\lambda(0)=1}$ and ${\lambda(r)\approx 0}$ when ${r\gg 0}$. We refer to any such $\lambda$ satisfying these properties as a \emph{blending function}. In our results, we choose:
\begin{equation}
    \lambda(h) = e^{-\beta\max\{0,h\}},
\end{equation}
with parameter ${\beta > 0}$, which satisfies the required properties of a blending function. The rationale behind the choice of the controller in~\eqref{eq:regulation-safety-filter} is that as $h_{I}(\bx)\rightarrow 0$,  we have $\bk(\bx,t)\rightarrow\bk_{\rm{b}}(\bx)$, a particular choice of input that ensures forward invariance of $\mathcal{S}$, whereas when $h_{I}(\bx)\gg 0$ we have $\bk(\bx,t)\rightarrow\bk_{\rm{d}}(\bx,t)$.
This means that the backup controller is used when the system is at the boundary of the safe set while the desired controller is used far inside the safe set.
The following theorem outlines the main properties of the safety filter in~\eqref{eq:regulation-safety-filter}.

\begin{theorem}[blended safety filter~\cite{singletary2022onboard}]
    Let the conditions of Theorem \ref{thm:implicit-safety} hold and suppose that ${\bk_{\rm{d}}\,:\,\mathcal{X}\times\R\rightarrow\mathcal{U}}$ is locally Lipschitz in its first argument and piecewise continuous in its second argument. Then, ${\bk\,:\,\mathcal{X}\times\R\rightarrow\mathcal{U}}$ as defined in~\eqref{eq:regulation-safety-filter} is locally Lipschitz in its first argument, piecewise continuous in its second argument, and enforces the forward invariance of $\mathcal{S}$.
\end{theorem}

When implementing Guardrails on the VISTA, we leverage the controller in~\eqref{eq:regulation-safety-filter}, where the desired controller $\bk_{\rm d}$ is a human pilot while the backup controller $\bk_{\rm b}$, the safety functions $h$, $h_{\rm b}$, and the underlying dynamical model are outlined shortly.
The blending strategy in~\eqref{eq:regulation-safety-filter} allowed us to smoothly transition from pilot inputs to backup commands when approaching the boundary of the safe operation domain, operate on the edge of this domain for significant durations of time, and return control authority to the pilot whenever its actions were deemed safe based on the safety function $h_{I}$ in~\eqref{eq:implicit-h}. Preliminary results on applying Guardrails to quadrotors \cite{DrewRAL22-geo,DrewIROS22}, rather than fixed-wing aircraft, are highlighted in ``Guardrails for High-Speed Geofencing on Quadrotors."

\begin{sidebar}{Guardrails for High-Speed Geofencing on Quadrotors}\label{sidebar:drone}

\setcounter{sequation}{0}
\renewcommand{\thesequation}{S\arabic{sequation}}
\setcounter{stable}{0}
\renewcommand{\thestable}{S\arabic{stable}}
\setcounter{sfigure}{0}
\renewcommand{\thesfigure}{S\arabic{sfigure}}

\sdbarinitial{P}reliminary steps toward applying our Guardrails framework on fixed-wing aircraft involved the application of Guardrails to high-speed geofencing with quadrotors \cite{DrewRAL22-geo,DrewIROS22}. Here, we model the quadrotor using a 13-dimensional state:
\begin{sequation}
    \bx = (\bp,\bq,\bv,\bomega),
\end{sequation}
where ${\bp\in\R^3}$ is the Cartesian position, ${\bq\in\mathbb{S}^3}$ is a unit quaternion describing the orientation, ${\bv\in\R^3}$ is the velocity, and ${\bomega\in\R^3}$ is the angular velocity. The dynamics of the quadrotor are modeled as a control affine system \eqref{eq:dyn} with:
\begin{sequation}
    \underbrace{
    \odv{}{t}
    \begin{bmatrix}
        \bp \\ \bq \\ \bv \\ \bomega
    \end{bmatrix}}_{\dot{\bx}}
    =
    \underbrace{
    \begin{bmatrix}
        \bv \\ \frac{1}{2}\bq \otimes \bomega_{\bq} \\ -g\be_{z} \\ - \bJ^{-1}\bomega \times \bJ \bomega
    \end{bmatrix}}_{\bf(\bx)}
    +
    \underbrace{
    \begin{bmatrix}
        \bzero_{3\times 1} & \bzero_{3\times 3} \\
        \bzero_{4 \times 1} & \bzero_{4\times 3} \\ 
        \frac{1}{m}\bR(\bq)\be_{z} & \bzero_{3 \times 3} \\ 
        \bzero_{3\times 1} & \bJ^{-1}
    \end{bmatrix}}_{\bg(\bx)}
    \underbrace{
    \begin{bmatrix}
        \tau \\ \bM
    \end{bmatrix}}_{\bu},
\end{sequation}
where the input $\bu=(\tau,\bM)\in\R^4$ consists of the thrust $\tau\in\R$ and moment $\bM\in\R^3$ generated by the propellers, $\bomega_{\bq}=(0,\bomega)$ is the pure quaternion representation of the angular velocity, $g$ is the acceleration due to gravity, $\be_z=(0,0,1)$ is the unit vector in the $z$ direction, $\bJ\in\R^{3\times 3}$ is the inertia matrix represented in the body frame, $m$ is the mass of the quadrotor, and $\bR(\bq)\in\mathrm{SO}(3)$ is the rotation matrix associated with $\bq$.

The main control objective is to keep the quadrotor's position within a box in 3D space (cf. Fig. \ref{sfig1}),
characterized by the safety constraint:
\begin{sequation}
    \begin{aligned}
        h(\bx) = & \min\{h_{x}(\bx),\,h_{y}(\bx),\,h_{z}(\bx)\}, \\ 
        h_{x}(\bx) = & r_{x}^2 - (p_{x} - x_{\rm c})^2, \\ 
        h_{y}(\bx) = & r_{y}^2 - (p_{y} - y_{\rm c})^2, \\ 
        h_{z}(\bx) = & r_{z}^2 - (p_{z} - z_{\rm c})^2,
    \end{aligned}
\end{sequation}
where $\bp=(p_x,p_y,p_z)$, which defines a box with side lengths $(r_x,r_y,r_z)$ centered at $(x_{\rm c},y_{\rm c},z_{\rm c})$. To generate an implicit control invariant set, we leverage a geometric tracking controller, similar to that in \cite{LeeCDC10}, as a backup controller that tracks reference velocity commands. Given a reference velocity command ${\bv_{\rm r}=(v_{x,{\rm r}},v_{y,{\rm r}},v_{z,{\rm r}})}$, the backup controller $\bk_{\rm b}$ ensures that ${\bv\rightarrow\bv_{\rm r}}$
where:
\begin{sequation}
    v_{i,{\rm r}} = \begin{cases}
        0 & h_i(\bx) \geq \delta_{i}, \\ 
        -(\delta_i - h_i(\bx)) & h_i(\bx) < \delta_{i},
    \end{cases}
\end{sequation}
for ${i \in \{x, y, z\}}$.
This backup maneuver attempts to bring the quadrotor to a zero velocity state at a position some positive distance from the boundary of the geofence. The backup set $\mathcal{C}_{\rm b}$ associated with this backup controller is governed by:
\begin{sequation}
    h_{\rm b}(\bx) = \min\{\epsilon - \|\bv\|, h(\bx)\}
\end{sequation}
for some $\epsilon>0$, describing states within the box with small velocity. This backup set is then used to generate an implicit control invariant set as in \eqref{eq:implicit-safe-set}, which was used to construct a blended safety filter as in \eqref{eq:regulation-safety-filter}, where the desired controller $\bk_{\rm{d}}(\bx,t)$ represents commands from a human pilot.

\sdbarfig{\includegraphics[width=19.0pc]{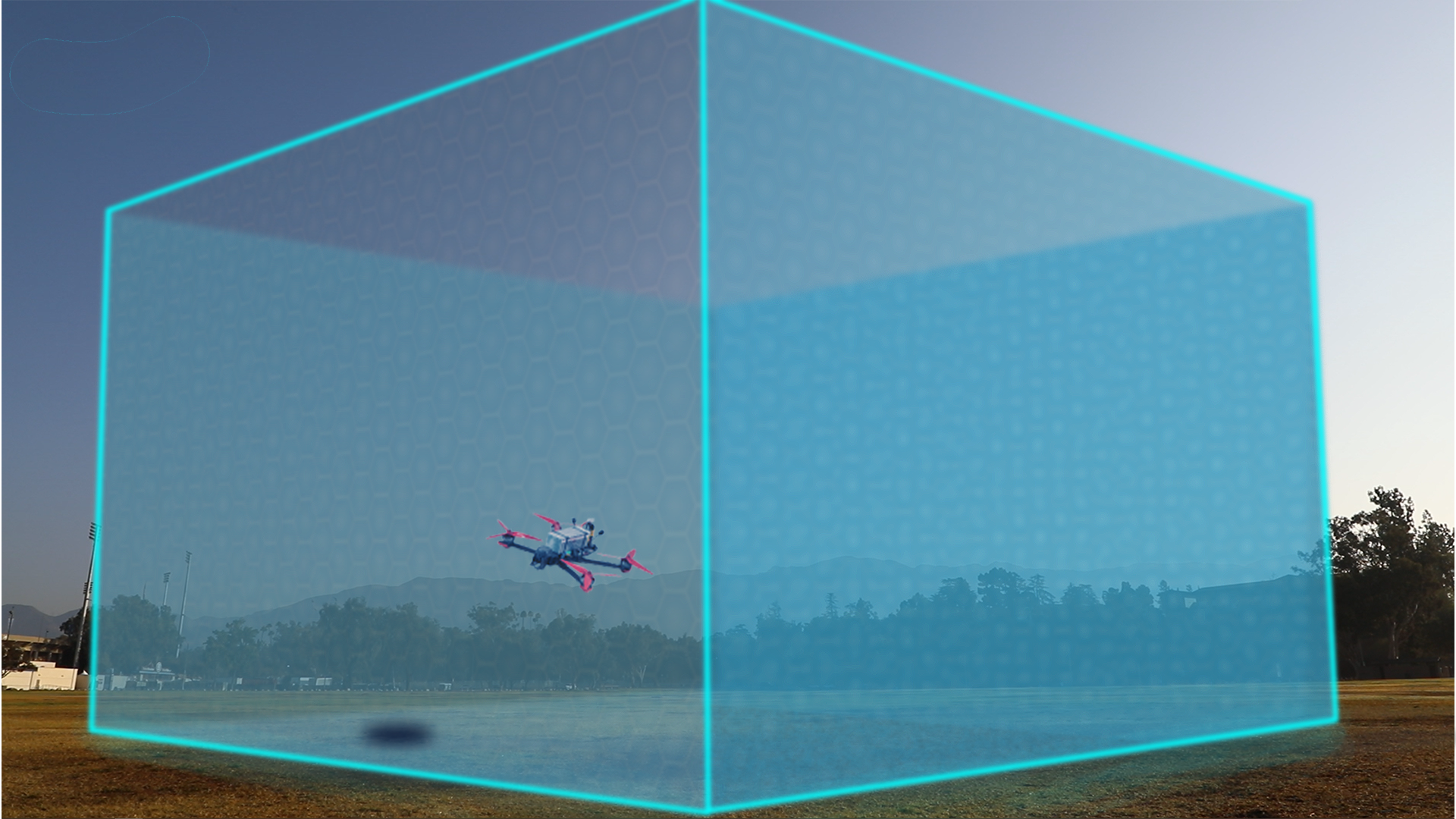}}{3D geofence from the application of Guardrails to a quadrotor. Figure adapted from \cite{DrewRAL22-geo}. \label{sfig1}}

The resulting safety filter was deployed in high-speed outdoor flight tests with the objective of avoiding flying through geofences.  A snapshot of these flight tests is highlighted in Fig. \ref{sfig2}, where the quadrotor is able to safely stop from a speed of over 100 km/h without crossing a vertical geofence.

\sdbarfig{\includegraphics[width=19.0pc]{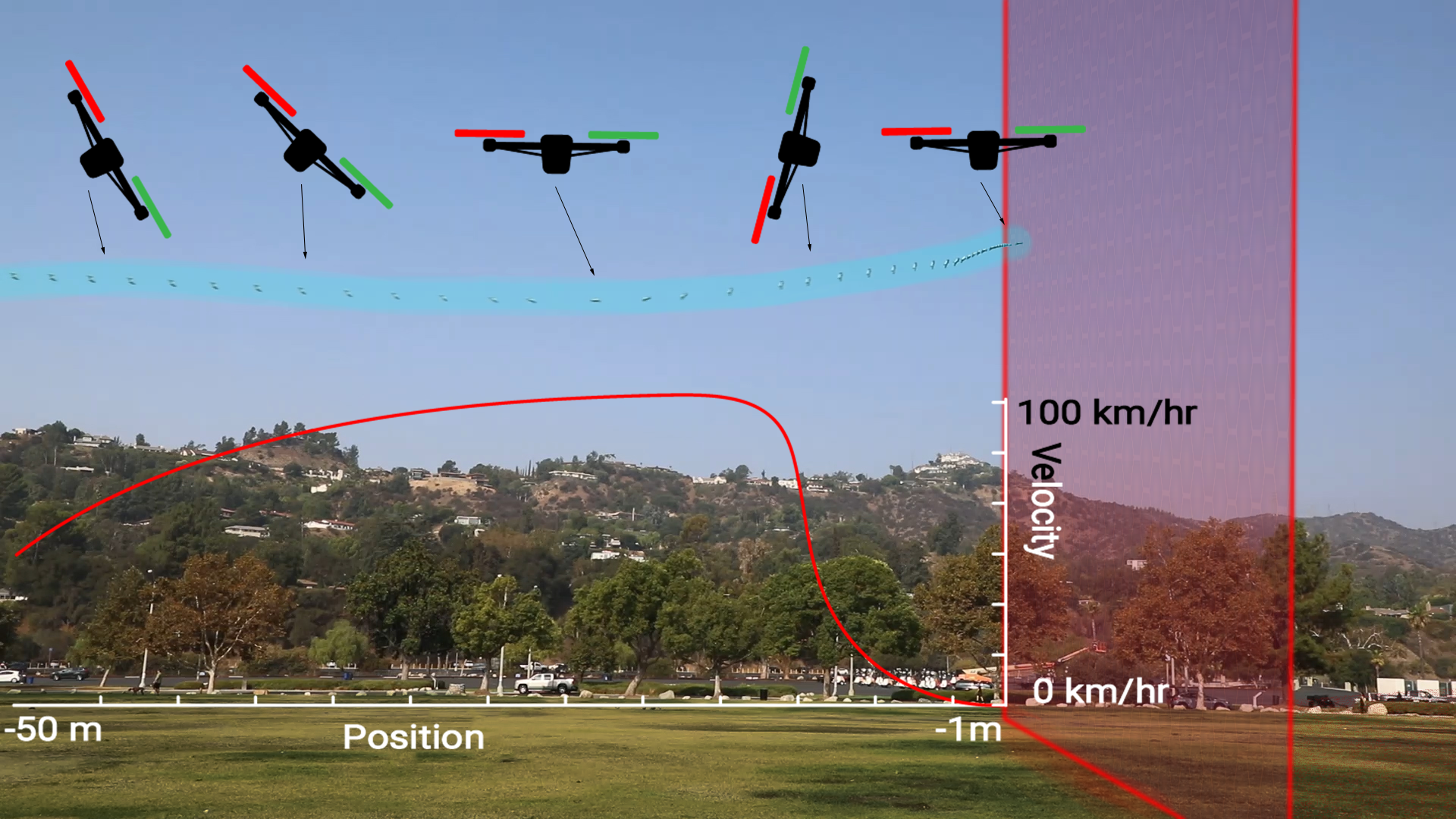}}{The implicit safe set and resulting blended safety filter ensures that a quadrotor does not cross a vertical geofence despite moving at high speeds. Figure adapted from \cite{DrewRAL22-geo}. \label{sfig2}}















\end{sidebar}

\section{Invariant Sets for Fixed-Wing Aircraft}

In this section we outline how the preceding technical developments may be applied to construct control invariant sets on fixed-wing aircraft. We first introduce the dynamic model used in our flight tests, describe the safety constraints considered, and finally discuss the various backup maneuvers used during testing.

\subsubsection{Dynamical Model}

To describe the aircraft's motion, we use a simplified dynamical model from~\cite{stephens2021realtime}, given by~(57)-(64) on page 57 of this reference.
This model was derived from the body-frame six-degree-of-freedom equations of motion in~\cite{stevens2016} with the following simplifying assumptions.
\begin{itemize}
    \item The aircraft is modeled as a point mass.
    \item The angle of attack and the angle of side slip are assumed to be zero.
    \item The roll mode (i.e., the roll control system of the aircraft) is modeled as a first-order system.
    \item Similarly, the load factor dynamics are also modeled as a first-order system.
    \item Gravity and load factor are the only accelerations on the aircraft.
\end{itemize}

The model describes the evolution of the Euler angles  $\phi$, $\theta$, $\psi$ (i.e., the roll, pitch, and yaw angles, respectively);
the north and east position coordinates $p_{\rm N}$ and $p_{\rm E}$, and the altitude $H$;
the roll rate $P$ (more precisely, the angular velocity about the forward axis of the aircraft);
the load factor $N_{z}$;
and the true airspeed $V_{\rm T}$.
By omitting the equation of the airspeed (because it is not being controlled by the Guardrails system in the experiments), and by neglecting wind disturbances, the dynamical equations are summarized as:
\begin{equation}\label{eq:model}
\begin{aligned}
    \dot{\phi} & = P + \frac{N_{z} g}{V_{\rm T}} \sin \phi \tan \theta, \\ 
    \dot{\theta} & = \frac{g}{V_{\rm T}} (N_{z} \cos \phi - \cos \theta), \\ 
    \dot{\psi} & = \frac{N_{z} g \sin \phi}{V_{\rm T} \cos \theta}, \\ 
    \dot{p}_{\rm N} & = V_{\rm T} \cos \theta \cos \psi, \\ 
    \dot{p}_{\rm E} & = V_{\rm T} \cos \theta \sin \psi, \\ 
    \dot{H} & = V_{\rm T} \sin \theta, \\ 
    \dot{P} & = \frac{1}{\tau_P} (u_P - P), \\ 
    \dot{N}_{z} & = \frac{1}{\tau_{z}} (u_{z} - N_{z}).
\end{aligned}
\end{equation}
Here,
$g$ is the gravitational acceleration;
$u_P$ is the commanded roll rate and $\tau_P$ is the roll-mode time constant;
$u_{z}$ is the commanded load factor and $\tau_{z}$ is the time constant of the load factor dynamics.
In this model, we refer to:
\begin{equation}
    \bx = (\phi, \theta, \psi, p_{\rm N}, p_{\rm E}, H, P, N_{z}) \in \R^{8}
\end{equation}
as the state of the system, whereas:
\begin{equation}
    \bu = (u_P, u_{z}) \in \R^2
\end{equation}
is the control input.

For the case of wing-level flight (with ${\phi=0}$), the following simplified version of the model can be used to describe the altitude and pitch dynamics only:
\begin{equation}\label{eq:model_simplified}
\begin{aligned}
    \dot{H} & = V_{\rm T} \sin \theta, \\
    \dot{\theta} & = \frac{g}{V_{\rm T}}(N_{z} - \cos \theta), \\
    \dot{N}_{z} & = \frac{1}{\tau_{z}} (u_{z} - N_{z}),
\end{aligned}
\end{equation}
with state ${\bx = (H,\theta,N_{z})\in\R^3}$ and input ${\bu = u_{z}\in\R}$.

\subsubsection{Safety Constraints}

The safety of the aircraft depends on its state $\bx$ as characterized by the constraint set $\mathcal{C}$ as in \eqref{eq:constraint-set} whereby
the aircraft satisfies the constraints at a state $\bx$
if:
\begin{equation}
    h(\bx) \geq 0,
\end{equation}
where ${h : \R^n \to \R}$ is a scalar-valued locally Lipschitz function.
During flight tests we consider three classes of constraints: i) those imposed on load factors (i.e., the $g$-force experienced by the pilot), ii) those imposed on the altitude, iii) those imposed on the overall position of the aircraft.

\textbf{Load Factor Constraints.} 
When enforcing load factor limits, the safety function becomes:
\begin{equation}
    h_1(\bx) = N_{z} - N_{z,{\rm min}},
\end{equation}
and:
\begin{equation}
    h_2(\bx) = N_{z,{\rm max}} - N_{z},
\end{equation}
where $N_{z,{\rm min}}$ and $N_{z,{\rm max}}$ denote the minimum and maximum load factors, respectively.
Note that since the first order system ${\dot{N}_{z} = (u_{z} - N_{z})/\tau_{z}}$ in~\eqref{eq:model} does not exhibit overshoots, enforcing ${N_{z,{\rm min}} \leq u_{z} \leq N_{z,{\rm max}}}$ leads to the satisfaction of ${N_{z,{\rm min}} \leq N_{z} \leq N_{z,{\rm max}}}$.
Therefore, simply clamping the input between $N_{z,{\rm min}}$ and $N_{z,{\rm max}}$ guarantees safety (${h_1(\bx) \geq 0}$ and ${h_2(\bx) \geq 0}$) w.r.t.~load factor limits.

\textbf{Altitude Constraints.} 
When imposing a flight floor at height $H_{\rm min}$ the safety function is defined as:
\begin{equation}
    h_3(\bx) = H - H_{\rm min}.
\end{equation}
Similarly, for a flight ceiling at height $H_{\rm max}$ the safety function is given by:
\begin{equation}
    h_4(\bx) = H_{\rm max} - H.
\end{equation}

\textbf{Geofencing Constraints.} 
In case of geofencing, safety is captured by:
\begin{equation}
    h_5(\bx) = \bn_{\rm g} \cdot (\bp - \bp_{\rm g}),
\end{equation}
where 
$\bp=(p_{\rm N}, p_{\rm E})$
is the position of the aircraft,
$\bp_{\rm g}$ is the position of the geofence, and
$\bn_{\rm g}$ is the normal vector of the geofence.
Here, $h$ represents the distance of the aircraft from the geofence, which was then converted into a time-to-collision measure based on the velocity to design safe controllers.

\textbf{Combining Constraints.} 
In general, the aircraft may be subject to a collection of safety constraints, each described by a safety function $h_i$ as above and indexed by ${i\in\{1,\dots,N\}}$, where $N$ is the number of constraints. This collection of constraints may be combined into a single constraint by taking the minimum among all constraints:
\begin{equation}
    h(x) = \min_{i} h_i(x).
\end{equation}
While CBF approaches typically require $h$ to be continuously differentiable, in the Guardrails framework we only require Lipschitz continuity, which is preserved when taking the minimum.

\subsubsection{Backup Maneuvers}
The safety constraints define a collection of states $\mathcal{C}$ as in \eqref{eq:constraint-set} that are deemed to within prescribed state limits. Yet, as noted earlier, this constraint set may not be control invariant -- it may be impossible to keep the system in this set for all time. In our application to fixed-wing aircraft, we address this using implicit control invariant sets, wherein a backup safety maneuver is used to efficiently compute control invariant sets contained within these state limits. In what follows, we outline backup strategies used to compute these invariant sets for the aircraft model in \eqref{eq:model}.

The backup strategy, represented as $\bk_{\rm b}$ in~\eqref{eq:dyn-backup}, is a coordinated 180-degree turn at a constant altitude.
This maneuver allows the satisfaction of geofence constraints by turning away from directions crossing the geofence boundary, while also satisfying altitude limits (floor and ceiling) by keeping the altitude constant.
As mentioned earlier, the load factor limits are enforced by clamping the load factor command.

The coordinated turn is designed to be a steady-state motion where the aircraft’s roll angle is a constant that determines the turning radius as well as the corresponding normal load command needed to maintain the altitude.
The direction of the turn (left or right) is chosen based on which direction results in a shorter turn.
The aircraft is assumed to be upright ($|\phi|<90$ degrees).

The steady-state motion is given by ${\phi(t) \equiv \phi^*}$, ${\theta(t) \equiv 0}$, and ${H(t) \equiv H^*}$ with a constant roll angle $\phi^*$, zero pitch angle, and a constant altitude $H^*$.
Based on the model in~\eqref{eq:model}, this leads to:
\begin{equation} \label{eq:equilibrium}
\begin{aligned}
    \dot{\phi} = 0, \quad
    \dot{\theta} = 0, \quad
    & \implies \quad
    P(t) \equiv 0, \quad 
    N_{z}(t) \equiv N_{z}^* = \frac{1}{\cos \phi^*}, \\ 
    & \implies \quad
    \dot{\psi} = \frac{g}{V_{\rm T}} \tan \phi^*.
\end{aligned}
\end{equation}
By choosing the steady-state roll angle $\phi^*$, one obtains the corresponding normal load factor $N_{z}^*$ and the turning radius ${R = V_{\rm T}/\dot{\psi} = V_{\rm T}^2/(g \tan \phi^*)}$.

The backup controller $\bk_{\rm b}$ is designed to execute this coordinated turn.
While the details of its exact implementation are omitted, a controller that could achieve a similar result is:
\begin{equation}
\begin{aligned}
    u_{P} & = k_{{\rm b},1}(\bx) = {\rm sat} \big( K_{\phi}(\bar{\phi} \!-\! \phi) \big), \quad
    \bar{\phi} = \min \big\{ \phi^*,K_{\psi}(\psi^* \!-\! \psi) \big\}, \\
    u_{z} & = k_{{\rm b},2}(\bx) = {\rm sat} \bigg( \frac{1}{\cos \phi} \big( 1 + K_{H} (H^* - H) - K_{\theta} \theta \big) \bigg),
\end{aligned}
\end{equation}
where ${\rm sat}$ indicates saturating (clamping) the inputs at their limits.
The control input $u_{P}$ involves a feedback term driving the roll angle $\phi$ to the desired value $\bar{\phi}$ using a gain $K_{\phi}$.
The desired roll angle $\bar{\phi}$ matches the equilibrium value $\phi^*$ during the most of the turn, but towards the end of the turn is reduced to $K_{\psi}(\psi^* - \psi)$, where $K_{\psi}$ is a gain and $\psi^*$ is the desired yaw angle corresponding to the completion of a 180-degree turn.
The control input $u_{z}$ involves the feedforward term ${1/\cos \phi}$ corresponding to $N_{z}^*$ in~\eqref{eq:equilibrium} as well as feedback terms driving the altitude $H$ to the equilibrium $H^*$ and the pitch angle $\theta$ to zero using the gains $K_{H}$ and $K_{\theta}$, respectively.
The parameters of the controller were tuned using extensive simulations and using feedback from human pilots about how the resulting behavior felt. Similar backup control strategies for fixed-wing aircraft are reported in \cite{DavidACC26}.

For the coordinated turn maneuver, the backup safe set (occurring in~\eqref{eq:backup-set}) is the set of states where a 180-degree turn is completed.
This can be described, for example, by:
\begin{equation}
    h_{\rm b}(\bx) = \pm (\psi - \psi^*),
\end{equation}
where the sign depends on whether the turn is to the left or to the right.

\begin{figure*}[!ht]
    \centerline{\includegraphics[scale=1]{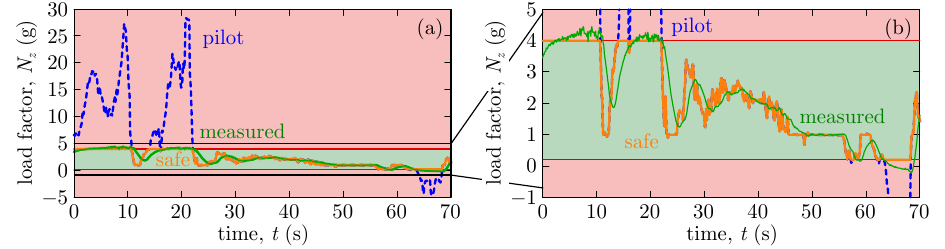}}
    \caption{Results associated with using Guardrails for load factor limiting. Here, the dashed blue curve represents the load factor requested by the pilot, the solid orange curve represents the safe load factor computed by Guardrails, and the thin green curve represents the measured load factor on the aircraft.
    Panel (b) is a zoom-in view of panel (a).}
    \label{fig:G-limits}
\end{figure*}

\section{Results}
In this section, we present the results from flight tests involving Guardrails as applied to a fixed-wing aircraft.
Before discussing the details of these tests, we first provide a brief overview of the results. Guardrails was extensively tested on the X-62 Variable In-Flight Simulation Test Aircraft (VISTA), a modified F-16 fighter jet developed by Lockheed Martin and used for testing advanced autonomy algorithms, in collaboration with the Edwards Air Force Base Test Pilot School. Guardrails completed all 14 test points, including pilot assault on safety limits. Multiple flights were conducted over the span of September 9 through September 20 of 2024 with 100\% success rate on these test points, consisting of combinations of the various constraints previously discussed. Video clips from a few of these tests can be viewed at \cite{GuardrailsVideo}.


\subsection{Experimental Setup: X-62 VISTA}
Although our Guardrails framework is applicable to general autonomous systems, our results in this paper stem from the application of Guardrails to the X-62 VISTA. Guardrails integrates into the VISTA's existing control stack through Lockheed Martin's System for Autonomous Control of the Simulation (SACS) module, which relays aircraft state information to the Guardrails module and converts control commands generated by Guardrails into lower-level actuator commands.

During our tests, Guardrails computes two different control inputs: commanded normal load factor ($u_{z}$) and commanded roll rate ($u_P$). The commanded normal load factor is converted into pitch rate commands by the SACS module, which are then passed to lower layers of the autonomy stack to determine the required surface deflections (elevator, aileron, rudder) to achieve the desired command. Roll rate commands are directly sent to lower layers of the autonomy stack and do not require conversion via the SACS module. While the VISTA is capable of being flown fully autonomously by an AI pilot, our tests are carried out with a human pilot in the loop. The pilot's inputs to the VISTA are pitch rate and roll rate commands. To interface these inputs with the Guardrails module, the pilot's pitch rate command is first converted to a normal load factor command. Then, Guardrails smoothly blends pilot inputs (${u_{P,{\rm d}}}$ and ${u_{z,{\rm d}}}$) with the backup controller to achieve safety of the overall semi-autonomous system. In what follows, we present extensive experimental results from a week of flight testing to illustrate the efficacy of Guardrails for achieving safety of highly dynamic autonomous systems. 

\subsection{Flight Tests}
We now present our key results obtained through in-flight enforcement of various safety constraints of increasing difficulty using Guardrails. All results were obtained in collaboration with the Edwards Air Force Base Test Pilot School between September 9 and September 20 of 2024.  

\begin{figure*}
    \centerline{\includegraphics[scale=1]{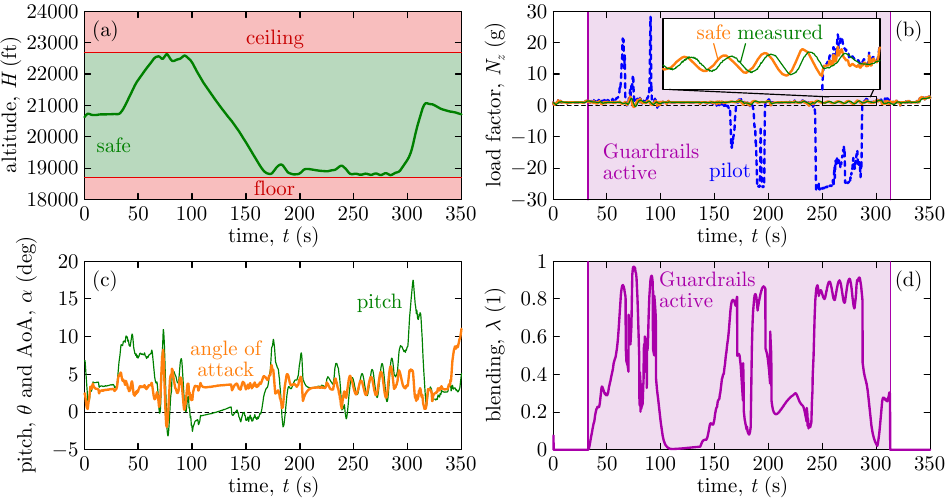}}
    \caption{Results from applying Guardrails for altitude limiting.
    In panel (a), the green curve denotes the evolution of the aircraft's altitude over time, with the green area denoting the safe region and the red area denoting the unsafe region. Panel (c) displays the evolution of the aircraft's pitch angle (thin green) and angle of attack (orange). Panel (b) illustrates the pilot's requested load factor (dashed blue), the commanded load factor computed by Guardrails (solid orange), and the measured load factor aboard the aircraft (thin green). Here, the purple region denotes the span of time where Guardrails is active (${\lambda>0}$), as quantified by the value of $\lambda$ in panel (d).}
    \label{fig:altitude_limits}
\end{figure*}

\subsubsection{Load Factor Limiting}
Our first results were obtained by leveraging Guardrails for limiting the load factor on the VISTA. The main objective of this test was to limit the load experienced by the pilot (i.e., how many g's they feel) to eliminate any pilot discomfort or safety trips. Limiting the load factor is critical to ensure the safety of a manned aircraft, as excessive loads can, in certain circumstances, lead to loss of consciousness by the pilot.

For this particular test, the load factor limit was set between 0.2 and 4 g's. Based on our chosen dynamical model of the aircraft in \eqref{eq:model}, the load factor $N_{z}$ is directly related to one of the control inputs $u_{z}$ through a first-order
actuator lag.
Given this close relationship between the constrained state of the system $N_{z}$ and the control input $u_{z}$, Guardrails achieves load factor limiting with an input saturation rather than a state limit, as noted earlier.

During this test, the pilot is instructed to pull up hard on the stick, requesting significantly more g's than allowed by the prescribed load factor limit. The results of applying Guardrails to ensure load factor limits are illustrated in Fig.~\ref{fig:G-limits} and the attached supplementary video \cite{GuardrailsVideo}. In Fig. \ref{fig:G-limits}, the dashed blue curve represents the load factor requested by the pilot (i.e., ${u_{z,{\rm d}}}$), the solid orange curve indicates the safe load factor computed by Guardrails (i.e., $u_{z}$), and the thin green curve shows the load factor experienced by the pilot as measured by sensors onboard the aircraft (i.e., $N_{z}$). As illustrated in Fig. \ref{fig:G-limits}, the load factor requested by the pilot far exceeds the limit while the commanded load factor computed by Guardrails always abides by the limit. Given that the commanded load factor, as computed by Guardrails, must pass through actuator dynamics before impacting the physical system, there is a slight delay between the commanded load factor (orange) and measured load factor (green curve) in Fig. \ref{fig:G-limits}, leading to small overshoots of the measured load factor beyond its limits. Despite these overshoots, the measured load factor stays well below the original, potentially unsafe load factor requested by the pilot and violates the imposed limits only by fractions of a g.

\subsubsection{Altitude Floor and Ceiling}
Our next test involves using Guardrails to impose limits on the altitude of the VISTA. Our objectives in this test were fourfold: i) ensure the altitude of the aircraft respects the prescribed limits; ii) modify the commanded load factor (the relevant control input for this test) smoothly and slowly so that no abrupt changes are felt by the pilot; iii) only restrict motion towards, not away from, the altitude limits; iv) always maintain partial control authority for the pilot. For this test, we placed a floor of 18,700 ft and a ceiling of 22,700 ft on the altitude of the VISTA. The initial altitude of the aircraft at the start of the test is just under 21,000 ft. During this test, the pilot is instructed to increase the aircraft's altitude from its initial value, attempting to drive the aircraft through the altitude ceiling for a period of time, after which the pilot repeats the same procedure for the altitude floor before returning the aircraft to its initial altitude.

The results of this altitude limiting test are provided in Fig. \ref{fig:altitude_limits}. 
As shown in Fig. \ref{fig:altitude_limits}(a), Guardrails ensures that the altitude limits are respected for all time, thereby achieving our first objective. That Guardrails achieves the second objective is illustrated in Fig. \ref{fig:altitude_limits}(b), which, like Fig. \ref{fig:G-limits}, shows the pilot's requested load factor $u_{z,{\rm d}}$, the commanded load factor $u_{z}$ computed by Guardrails, and the measured load factor $N_{z}$ aboard the aircraft. 
The satisfaction of objectives (iii) and (iv) can be inferred from Fig. \ref{fig:altitude_limits}(d), which portrays the evolution of $\lambda$, the parameter that blends inputs from the pilot and from the Guardrails, with $\lambda=0$ implying the pilot is in full control and $\lambda=1$ implying that Guardrails is in complete control. During this test, $\lambda$ remains strictly less than one for all time, indicating that the pilot always has partial control over the aircraft, thereby achieving objective (iv). The evolution of $\lambda$ also indicates that motion away from the altitude limits is not restricted: at around 100 seconds, when the pilot begins pitching the aircraft down, $\lambda$ quickly decreases to zero, allowing the pilot to freely move the aircraft away from the boundary. A similar phenomenon is seen just before the 300-second mark, where the aircraft is pitched back up off the lower altitude limit and $\lambda$ quickly decays to a low value.



\begin{figure*}
    \centerline{\includegraphics[scale=1]{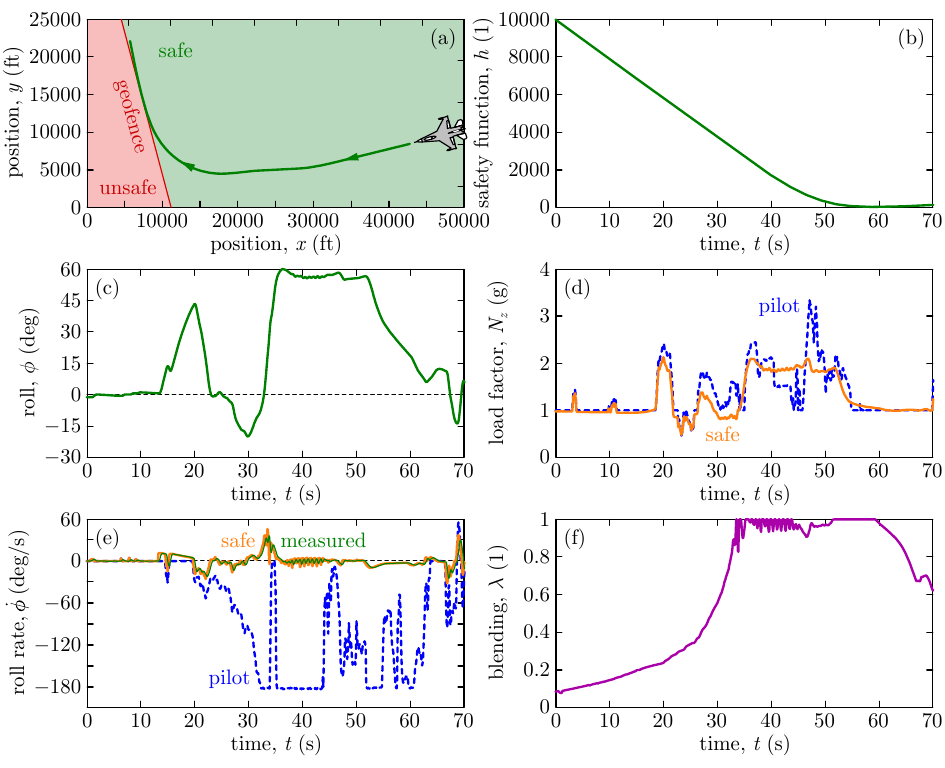}}
    \caption{Results from employing Guardrails for geofencing.
    Panel (a) highlights that the aircraft's trajectory (green curve) stays in the safe region (green area), avoiding the restricted airspace (red area).
    Panel (b) plots the safety function $h$ (a normalized time to collision measure associated with the distance to the geofence), whose positive value indicates safety.
    Panel (c) depicts how the aircraft rolls to abide by the geofence.
    Panels (d) and (e), respectively, show that the load factor and roll rate commanded by the pilot (dashed blue) are modified to a safe command by Guardrails (solid orange) which is tracked by the aircraft (thin green).
    The control authority of Guardrails is shown in panel (f).}
    \label{fig:geofencing}
\end{figure*}

\begin{figure*}
    \centerline{\includegraphics[scale=1]{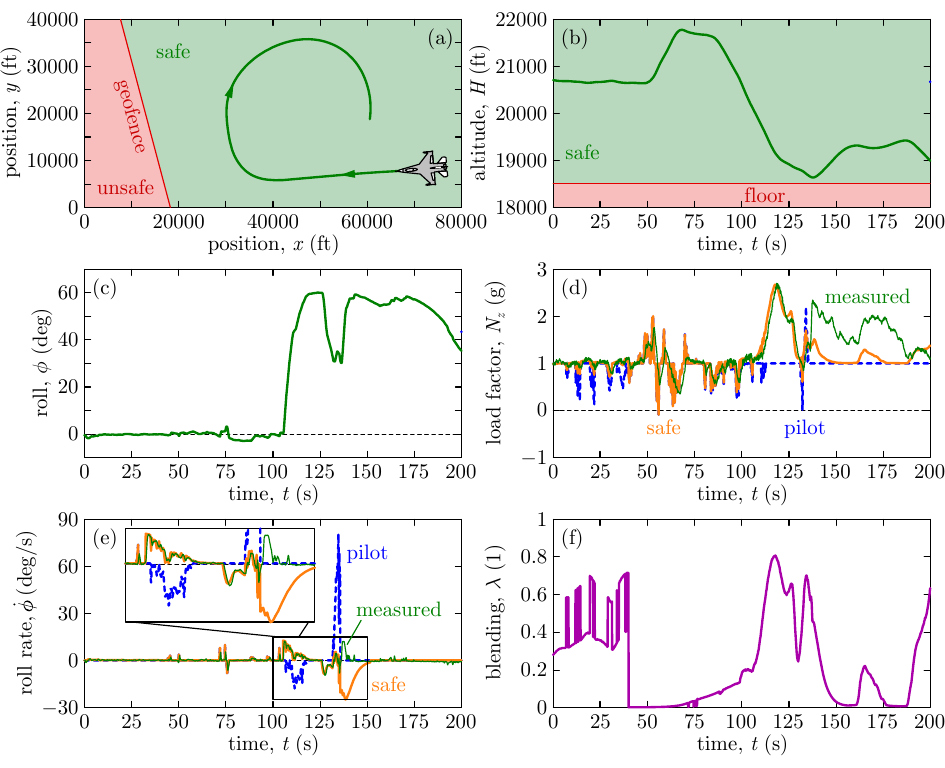}}
    \caption{Results from using Guardrails for simultaneous geofencing and altitude limiting.
    In panels (a) and (b), respectively, the VISTA's trajectory and altitude (green curve) are shown, with the safe (green) and unsafe (red) regions indicated.
    Panel (c) depicts the evolution of the roll angle.
    Panels (d) and (e) display the corresponding load factor and roll rate, respectively, as requested by the pilot (dashed blue), computed by Guardrails (solid orange), and measured aboard the aircraft (thin green).
    The partial control authority of Guardrails is quantified by the blending parameter $\lambda$ in panel (f).}
    \label{fig:geofence_and_floor}
\end{figure*}

\subsubsection{Geofencing}

Our next test leverages Guardrails for geofencing, wherein the objective is to keep the aircraft in a prespecified airspace; see Fig. \ref{fig:geofencing}. For this test, the aircraft begins with its heading perpendicular to the geofence (see Fig. \ref{fig:geofencing}(a)) and must perform a smooth coordinated turn to avoid flying through the geofence. Unlike the previous tests, performing this maneuver requires coordinating two different control inputs -- the commanded load factor, as in the previous tests, and the commanded roll rate.  Importantly, this maneuver must be performed while respecting hard limits on these control inputs. The commanded roll rate $u_P$ is related to the actual roll rate $P$ through a first-order actuator lag; hence, as discussed in the previous results, we enforce constraints on the roll rate through control limits rather than state limits.

During this test, the pilot is initially instructed to fly the aircraft toward the geofence. As the aircraft approaches the geofence (see Fig. \ref{fig:geofencing}(a)) and the distance to the geofence decreases (see Fig. \ref{fig:geofencing}(b)), Guardrails is gradually given more control authority over the aircraft, as indicated by the increasing value of $\lambda$ in Fig. \ref{fig:geofencing}(f). At around ${t \approx 25}$ seconds, the commands from the pilot and those computed by Guardrails begin to deviate significantly (Fig. \ref{fig:geofencing}(d) and Fig. \ref{fig:geofencing}(e)), with Guardrails issuing commands that cause the aircraft to bank and align its heading to be parallel with the geofence (see Fig. \ref{fig:geofencing}(a)). Once the aircraft is moving parallel to the geofence, the pilot is instructed to attempt to drive the aircraft into the geofence by issuing aggressive roll rate commands toward the unsafe region from ${t\approx 35}$ to ${t\approx 60}$ seconds, as illustrated in Fig. \ref{fig:geofencing}(e). Over this time span, Guardrails counteracts these unsafe commands by computing much smaller roll rate commands (Fig. \ref{fig:geofencing}(e)). Coupled with the fact that ${\lambda\approx 1}$ over this interval, this ensures that the actual roll rate measured onboard the aircraft is small, keeping the aircraft aligned with the geofence and preventing safety violations. 

\subsubsection{Geofencing and Altitude Limiting}
Our next test combines geofence and altitude floor limits, testing the ability of Guardrails to handle multiple state limits. During the test, the pilot is instructed to fly the aircraft directly at the geofence with a downward pitch of about five degrees, so that the aircraft comes close to both the geofence and altitude floor limits simultaneously. 


The results of this test are shown in Fig. \ref{fig:geofence_and_floor} and the attached supplementary video \cite{GuardrailsVideo}.
The test begins with the aircraft at level flight heading directly toward the geofence. At around ${t \approx 50}$ seconds, the aircraft pitches up to achieve an altitude of just under 22,000 ft before pitching down in an attempt to arrive at the altitude floor limit at a similar time to the geofence limit (see Fig. \ref{fig:geofence_and_floor}(b)). At around ${t \approx 100}$ seconds, Guardrails begins to intervene, as indicated by the value of $\lambda$ in Fig. \ref{fig:geofence_and_floor}(f), and banks the aircraft away from the geofence (Fig. \ref{fig:geofence_and_floor}(a)) by commanding positive roll rates (Fig. \ref{fig:geofence_and_floor}(e)), which causes the aircraft's roll angle to increase to about 60 degrees. Simultaneously, Guardrails produces load factor commands between 1-3 g's (Fig. \ref{fig:geofence_and_floor}(d)), causing the aircraft to pitch up so that it does not violate the altitude floor (Fig. \ref{fig:geofence_and_floor}(b)). 
At around ${t \approx 140}$ seconds, Guardrails is disengaged through a pilot overtake as indicated by the significant deviation between the measured load factor (thin green) and commanded load factor (solid orange) in Fig.~\ref{fig:geofence_and_floor}(d), concluding the experiment.

\begin{figure*}
    \centerline{\includegraphics[scale=1]{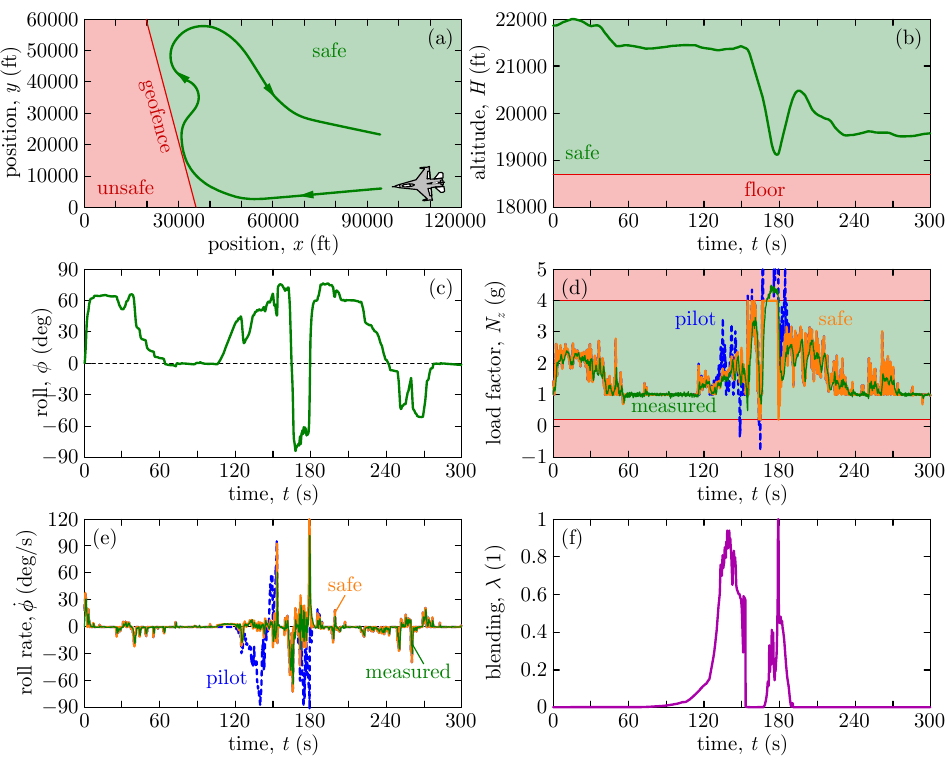}}
    \caption{Results from applying Guardrails to enforce combinations of geofence, altitude, and load factor constraints.
    The same information is shown with the same notations as in Fig. \ref{fig:geofence_and_floor}, while panel (d) also displays the safe (green shading) and unsafe load factor ranges (red shading).}
    \label{fig:floor_geo_Nz}
\end{figure*}

\subsubsection{Geofencing, Altitude, and Load Factor Limiting}

Our final tests employ Guardrails to enforce geofence constraints, altitude limits, and load factor limits simultaneously.
We show two test scenarios for this case, to demonstrate the capabilities of Guardrails in assuring safety in various situations.

In the first test, shown in Fig. \ref{fig:floor_geo_Nz}, the pilot is instructed to fly towards the geofence (Fig. \ref{fig:floor_geo_Nz}(a)) while descending from $22,000$ ft to approach the altitude floor (Fig. \ref{fig:floor_geo_Nz}(b)).
At ${t \approx 100}$ seconds, Guardrails starts to gradually take control (Fig. \ref{fig:floor_geo_Nz}(f)) and slowly bank the jet to a $75$ degree roll angle (Fig. \ref{fig:floor_geo_Nz}(c)), making it fly parallel to the geofence  (Fig. \ref{fig:floor_geo_Nz}(a)).
During this process, between ${t \approx 120}$ and ${t \approx 150}$ seconds, the pilot actively tries to bank towards the geofence but Guardrails prevents it from doing so (Fig. \ref{fig:floor_geo_Nz}(e)).
Afterwards, the pilot decides to fly the aircraft away from the geofence, and Guardrails gives the control authority back to the pilot at ${t \approx 150}$ seconds (Fig. \ref{fig:floor_geo_Nz}(f)).
Then, the pilot banks the aircraft in the other direction, decreasing the roll angle to $-75$ degrees, and turns the VISTA hard toward the geofence to assault it once again, while bringing the aircraft's nose down.
Guardrails intervenes for the second time, at around ${t \approx 180}$ seconds, to prevent the aircraft from entering the restricted airspace despite the pilot's commands.
At the same time, the aircraft arrives at the altitude floor (Fig. \ref{fig:floor_geo_Nz}(b)), and the pilot commands a large positive load factor to pull it up (Fig. \ref{fig:floor_geo_Nz}(d)).
In response, Guardrails allows the aircraft to pull up and avoid the altitude floor, but it limits the load factor to 4 g's to maintain the safety of the pilot.
Finally, control authority is given back to the pilot as it flies away from the geofence at a safe altitude.

\begin{figure*}
    \centerline{\includegraphics[scale=1]{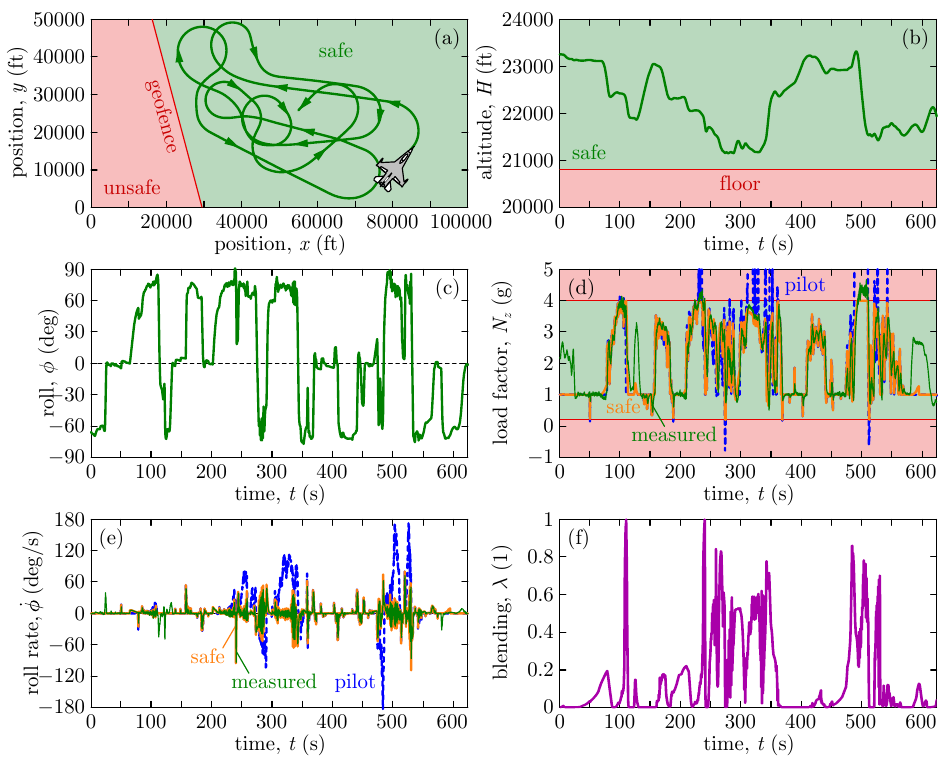}}
    \caption{Results of repeatedly using Guardrails to enforce combinations of geofence, altitude, and load factor constraints.
    The same information is shown with the same notations as in Fig. \ref{fig:floor_geo_Nz}.}
    \label{fig:floor_geo_Nz_multi}
\end{figure*}

In the second test, shown in Fig. \ref{fig:floor_geo_Nz_multi}, the combination of a geofence constraint, an altitude floor, and load factor limits is enforced, while the pilot is instructed to attempt the violation of these constraints several times during a $10$-minute experiment.
In each case, Guardrails successfully maintains the geofence constraint (Fig. \ref{fig:floor_geo_Nz_multi}(a)), even when it requires a roll action opposite to what is being commanded by the pilot (Fig. \ref{fig:floor_geo_Nz_multi}(e)).
Guadrails also keeps the altitude above the floor for all time (Fig. \ref{fig:floor_geo_Nz_multi}(b)) with the appropriate load factor command (Fig. \ref{fig:floor_geo_Nz_multi}(d)).
The load factor computed by Guardrails is within the prescribed safe limits, even when the pilot applies a hard pull on the stick, while the load factor measured on board exhibits slight overshoots beyond the limits.
To achieve safe behavior, Guardrails repeatedly claims back control authority when needed to guarantee safety (Fig. \ref{fig:floor_geo_Nz_multi}(f)), and gives it back to the pilot otherwise.
Through smoothly blending pilot commands with safe actions, Guardrails enables the jet to be operated with formal guarantees of safety, including flights at the edge of the safe operation domain bounded by multiple constraints.

\section{Discussion}

\subsection{Takeaways from Flight Testing}

Overall, Guardrails showed success in enforcing g-limits, altitude bounds, geofence constraints, and their combination in a variety of scenarios.
Throughout the flight tests, Guardrails managed to supervise a human pilot aboard the F-16 jet and establish safe behaviors by smoothly blending the pilot actions with backup strategies that provided provably safe normal load and roll rate commands based on a reduced-order model of the aircraft dynamics.
The pilot actions were directly executed whenever the situation was deemed safe based on the corresponding safety function, and otherwise Guardrails gradually took the control authority away from the pilot, allowing the satisfaction of safety constraints even in the presence of adversarial, intentionally unsafe pilot actions.
Importantly, Guardrails enabled operation at the safety limit, such as flight along geofence boundaries or at an altitude ceiling or floor, for sustained durations of time. 
A discussion on the observed minor safety violations as well as on limitations and potential future improvements is given below.

\subsection{Design Decisions}

We now summarize some of the important design decisions that were made to facilitate the successful implementation of Guardrails during real-world flight testing.
These decisions seek a balance between minimizing safety violations, reducing the complexity of the control design, while maximizing the control authority of the pilot.

\subsubsection{Input vs. State Limits}
As noted in our results section, one of our control inputs is the commanded load factor $u_{z}$, which is related to the actual normal load factor $N_{z}$ through a first order actuator model. When faced with tests that required limiting the load factor, we chose to simply limit the commanded load factor by clamping the input at the specified bounds, rather than constructing a control invariant set to enforce the corresponding state limit. This greatly simplifies the control design, but may result in small violations of the prescribed state limit as illustrated in Fig. \ref{fig:G-limits}(b),
Fig. \ref{fig:floor_geo_Nz}(d), and Fig. \ref{fig:floor_geo_Nz_multi}(d), where the measured state $N_{z}$ exhibits small violations of the required limit despite the fact that the commanded input $u_{z}$ remains within the limits. These violations occur because, while we represent the actuator dynamics using a first-order lag model, in reality these actuator dynamics are much more complicated. This discrepancy between our model and the true dynamics of the system causes the actual $N_{z}$ (labeled as ``measured" in the plots) to overshoot the commanded signal. These violations could be eliminated by either adding an additional margin to the limit enforced on the commanded load factor or by treating the $N_{z}$ limits as a state limit rather than an input limit, and including a more complex actuator model. Further justification for this simplified choice of actuator model is expanded on in the the following subsection.

\subsubsection{Simplified Models for Control Design}
Aircraft dynamics are inherently complex, high-dimensional, nonlinear, and are subject to uncertain parameters and disturbances, making the use of high-fidelity aircraft models challenging for control design. To overcome this, aerospace engineers typically employ highly simplified models for control design, trimmed at certain operating conditions, which provide local approximations of the full-order aircraft dynamics \cite{lavretsky2012robust,stevens2016}. Guardrails follows a similar philosophy by leveraging a simplified, reduced-order model of an aircraft that captures the essential features (kinematics of the pitch and roll axes) relevant to the safety constraints in our tests. Higher order dynamics are abstracted away by assuming a first-order actuator model for our two control inputs, the roll rate and normal load factor.

It is essential to note that the safety assurances offered by Guardrails are predicated on the choice of model used for control design, raising the question if the simplified models used in our results are accurate enough to establish safety guarantees. Fortunately, in the context of robotic systems, there is a large body of supporting work on using CBFs for reduced-order models while still making practical guarantees on safety of the full-order system~\cite{molnar2022modelfree, TamasACC23,cohen2024reduced,CohenACC25}. These approaches have been successfully deployed to ensure safety on highly dynamic robotic systems, such as drones, quadrupeds, and hopping robots, despite using dramatically simplified models for the control design, such as single integrators and unicycles. These well-established results apply to our flight control case study as well --- despite abstracting away high-order terms in the aircraft dynamics, the theoretical results from, e.g., \cite{molnar2022modelfree, TamasACC23,cohen2024reduced,CohenACC25}, may be employed to provide practical safety assurances for the original full-order dynamics of the aircraft.

\subsubsection{Reduced Pilot Authority}
The performance of the Guardrails system is determined by the interplay of the pilot commands and the safe backup inputs as well as by how the pilot authority is reduced in safety-critical situations.
This interplay can be observed, for example, in our altitude limiting tests (see Fig. \ref{fig:altitude_limits}(a)), where the aircraft was bouncing up and down with mild oscillatory motion when it was near the altitude limits and the pilot was attempting to push the aircraft through these limits (between ${t \approx 250}$ and ${t \approx 300}$ seconds). The oscillatory motion is due to the fluctuations in $\lambda$ (see Fig. \ref{fig:altitude_limits}(d)), causing Guardrails to rapidly transition between the commanded pilot inputs and the push-back from the backup controller to prevent safety violation. These oscillations could be reduced by choosing a more aggressive blending function $\lambda$, which would reduce the pilot's control authority more significantly upon approaching the constraint boundary and allow the backup controller to (almost) independently control the aircraft. Alternatively, one could choose a more aggressive backup maneuver to achieve a similar result. In our testing, we chose not to implement these changes as ensuring adequate pilot authority near the constraint boundary was deemed a high priority. Future efforts will focus on mitigating oscillations while retaining some level of control authority for the pilot. Preliminary efforts towards this goal are presented in \cite{DavidACC26}.

\subsection{Limitations}

While Guardrails successfully enforced safety across various scenarios through extensive real-world flight testing, there are details in the design of Guardrails that could be further investigated and improved to enhance performance and optimize overall system behavior on the VISTA and other autonomous systems.
These include making Guardrails less conservative so that the space where the pilot has significant control authority is maximized.
For example, we observed some conservativeness when the aircraft approached the altitude floor constraint as Guardrails activated relatively far from the boundary.
Less conservative behavior could be achieved through optimizing the backup maneuvers, or by fine-tuning the blending function $\lambda$.
Tuning the blending function could also help mitigate the oscillations of the aircraft when the pilot is pushing against the floor --- this, however, requires reducing the control authority of the pilot, marking a trade-off in the control design.
Furthermore, our backup maneuvers could be improved, for example, by formulating the desired bank angle as a function of the distance from the geofence, which could make the transition into the geofence bank more gradual.

\section{CONCLUSION}
In this article, we introduced Guardrails, a framework for safety filtering on highly dynamic autonomous systems. Guardrails is rooted in the theory control barrier functions and enables the construction of control invariant sets and computationally efficient safety filters for high-dimensional nonlinear systems. We implemented Guardrails on the VISTA -- a modified F-16 fighter jet -- conducting numerous real-world flight tests to ensure the satisfaction of safety-critical constraints such as g-force limits, altitude limits, and geofence constraints.

There are several directions for future work to expand the capabilities of Guardrails on the VISTA.
These include enforcing additional safety constraints on a variety of states.
For example, angle of attack and pitch angle limits may help prevent rapid speed losses that were experienced during some of the tests.
Similarly, one could establish collision avoidance with nearby aircraft by synthesizing a backup controller that executes an evasive maneuver and incorporating the distance of the two aircraft into the safety function. Other directions include leveraging gradient information of the safety function in a computationally efficient manner to further improve performance.

While our results in this paper focused on the application of Guardrails to the VISTA in the context of safety-critical flight control, we stress that our overall framework is broadly applicable to a wide range of autonomous systems, such as ground vehicles and legged robots.
By using Guardrails as an add-on safety module for these systems, novel decision making, motion planning, and control algorithms can be tested and deployed without additional safety risks, which greatly facilitates the development of next-generation autonomous systems.

\section{ACKNOWLEDGMENT}

The authors thank Al Moser for his help and support in this research.
This research was supported by the AFOSR Test and Evaluation program, grant FA9550-22-1-0333.

\section{Author Information}


\begin{IEEEbiography}{{A}ndrew W. Singletary}{\,} 
(asingletary@3lawsrobotics.com) received the B.S. degrees in mechanical engineering and nuclear and radiological engineering from the Georgia Institute of Technology in 2017, and the M.S. and Ph.D. degrees in mechanical engineering from the California Institute of Technology in 2019 and 2022, respectively. His doctoral work at Caltech focused on safety-critical control for robotics, including control barrier functions and real-time safety filtering frameworks. He is the co-founder and CEO of 3Laws, where he leads the translation of these control-theoretic advances into deployable safety technologies for autonomous robotic systems. He was named to the Forbes 30 Under 30 Science list in 2024.
\end{IEEEbiography}

\begin{IEEEbiography}{{M}ax H. Cohen}{\,}(mhcohen2@ncsu.edu) is an Assistant Professor of Electrical and Computer Engineering at North Carolina State University. Prior to joining NC State, he served as a Postdoctoral Scholar at the California Institute of Technology from 2023-2025. He earned the B.S. in Mechanical Engineering from the University of Florida in 2018, the M.S. in Mechanical Engineering from Boston University in 2022, and the Ph.D. in Mechanical Engineering from Boston University in 2023. He was awarded an NSF Graduate Research Fellowship in 2019 and the best paper award at the 2025 Conference on Learning for Dynamics and Control. His research interests include nonlinear control and learning-based control with applications to robotics and autonomous systems.
\end{IEEEbiography}

\begin{IEEEbiography}{{T}amas G. Molnar}{\,}(tamas.molnar@wichita.edu) is an Assistant Professor of Mechanical Engineering at the Wichita State University since 2023. Beforehand, he held postdoctoral positions at the California Institute of Technology, from 2020 to 2023, and at the University of Michigan, Ann Arbor, from 2018 to 2020. He received the Ph.D. and M.S. degrees in Mechanical Engineering and the B.S. degree in Mechatronics Engineering from the Budapest University of Technology and Economics, Hungary, in 2018, 2015, and 2013. His research interests include nonlinear dynamics and control, safety-critical control, and time delay systems with applications to connected automated vehicles, robotic systems, and autonomous systems.
\end{IEEEbiography}

\begin{IEEEbiography}{{A}aron D. Ames}{\,}(ames@caltech.edu) received the B.S. degree in mechanical engineering and the B.A. degree in mathematics from the University of St. Thomas, Saint Paul, MN, USA, in 2001, the M.A. degree in mathematics and the Ph.D. degree in electrical engineering and computer sciences from UC Berkeley, Berkeley, CA, USA, in 2006.
He began his faculty career with Texas A\&M University, in 2008. He was an Associate Professor of mechanical engineering and electrical \& computer engineering with the Georgia Institute of Technology and a Postdoctoral Scholar of control and dynamical systems with Caltech from 2006 to 2008. He is currently the Bren Professor of mechanical and civil engineering and control and dynamical systems with the California Institute of Technology, Pasadena, CA, USA.
His research interests include in the areas of robotics, nonlinear control, and hybrid systems, with a special
focus on applications to bipedal robotic walking---both formally and through experimental validation.
Dr. Ames was the recipient of the 2005 Leon O. Chua Award for achievement in nonlinear science at UC Berkeley, 2006 Bernard Friedman Memorial Prize in Applied Mathematics, NSF Career Award in 2010, Donald P. Eckman Award in 2015, and the 2019 Antonio Ruberti Young Researcher Prize.
\end{IEEEbiography}





\bibliography{guardrails}
\bibliographystyle{IEEEtran}

\endarticle

\end{document}